\documentclass[5p,sort&compress]{elsarticle}

\usepackage{lineno,hyperref}
\modulolinenumbers[5]

\journal{Information Sciences}

\usepackage{graphicx}
\usepackage{amsmath,amssymb,amsfonts}
\usepackage{amsthm}
\usepackage{algorithmic}
\usepackage{color}
\usepackage[toc,page]{appendix}
\usepackage{environ}
\usepackage{tikz}
\usetikzlibrary{shapes,arrows}
\usepackage{natbib}
\usepackage{color}
\usepackage{rotate}

\newcommand{\bea}{\begin{eqnarray*}}
\newcommand{\eea}{\end{eqnarray*}}
\newcommand{\bne}{\begin{equation*}}
\newtheorem{example}{Example}
\newcommand{\ede}{\end{equation*}}

\newcommand{\bnen}{\begin{equation}}
\newcommand{\eden}{\end{equation}}
\newcommand{\bean}{\begin{eqnarray}}
\newcommand{\eean}{\end{eqnarray}}
\newcommand{\bnsn}{\begin{subequations}}
\newcommand{\edsn}{\end{subequations}}

\newcommand{\bna}{\begin{array}}
\newcommand{\eda}{\end{array}}
\newcommand{\bnm}{\begin{enumerate}}
\newcommand{\edm}{\end{enumerate}}
\newcommand{\bni}{\begin{itemize}}
\newcommand{\edi}{\end{itemize}}

\newtheorem{thm}{Definition}
\newtheorem{rem}{Remark}
\newtheorem{pro}{Proposition}

\begin{document}
	
	\begin{frontmatter}
		
		\title{Interpretable neural networks based on continuous-valued logic and multicriteria decision operators		}
	
		\cortext[mycorrespondingauthor]{Corresponding author}
		
		\author[1,2]{Orsolya Csisz\'ar\corref{mycorrespondingauthor}}
		\ead{orsolya.csiszar@nik.uni-obuda.hu}
	
		\author[3]{G\'abor Csisz\'ar}
		\ead{Gabor.Csiszar@mp.imw.uni-stuttgart.de}
	
		\author[4]{J\'ozsef Dombi}
		\ead{dombi@inf.u-szeged.hu}
		
		\address[1]{Faculty of Basic Sciences, University of Applied Sciences Esslingen, Esslingen, Germany}
		\address[2]{Institute of Applied Mathematics, \'Obuda University, Budapest, Hungary}
		\address[3]{Institute of Materials Physics, University of Stuttgart, Stuttgart, Germany}
		\address[4]{Institute of Informatics, University of Szeged, Szeged, Hungary}
				
	\begin{abstract}
		Combining neural networks with continuous logic and multicriteria decision making tools can reduce the black box nature of neural models. In this study, we show that nilpotent logical systems offer an appropriate mathematical framework for a hybridization of continuous nilpotent logic and neural models, helping to improve the interpretability and safety of machine learning. 
		In our concept, perceptrons model soft inequalities; namely membership functions and continuous logical operators. We design the network architecture before training, using continuous logical operators and multicriteria decision tools with given weights working in the hidden layers.  Designing the structure appropriately leads to a drastic reduction in the number of parameters to be learned. The theoretical basis offers a straightforward choice of activation functions (the cutting function or its differentiable approximation, the squashing function), and also suggests an explanation to the great success of the rectified linear unit (ReLU). 
		In this study, we focus on the architecture of a hybrid model and introduce the building blocks for future application in deep neural networks. 
		The concept is illustrated with some toy examples taken from an extended version of the tensorflow playground.
		
	\end{abstract}
		
		
	\end{frontmatter}

\section{Introduction}

AI techniques, especially deep learning models are revolutionizing the business and technology world.  One of the greatest challenges is the increasing need to address the problem of interpretability and to improve model transparency, performance and safety. Although deep neural networks have achieved impressive experimental results e.g. in image classification, they may surprisingly be unstable when it comes to adversarial perturbations, that is, minimal changes to the input image that cause the network to misclassify it \cite{adv1, adv2, adv3, adv}. Combining deep neural networks with structured logical rules and multicriteria decision tools, where logical operators are applied on clusters created in the first layer, contributes to the reduction of the black box nature of neural models.  Aiming at interpretability, transparency and safety, implementing continuous-valued logical operators offers a promising direction. 

Although boolean units and multilayer perceptrons have a long history, to the best of our knowledge there has been little attempt to combine neural networks with continuous logical systems so far.
The basic idea of continuous logic is the replacement of the space of truth values $\{T,F\}$ by a compact interval such as $[0, 1]$.  This means that the inputs and the outputs of the extended logical gates are real numbers of the unit interval, representing truth values of inequalities. Quantifiers $\forall x$  and $\exists x$ are replaced by $\sup_x$ and $\inf_x$, and logical connectives are continuous functions. Based on this idea, human thinking and natural language can be modeled in a sophisticated way. 

Among other families of many-valued logics, t-norm fuzzy logics are broadly used in applied fuzzy logic and fuzzy set theory as a theoretical basis for approximate reasoning. In fuzzy logic, the membership function of a fuzzy set represents the degree of truth as a generalization of the indicator function in classical sets.
Both propositional and first-order (or higher-order) t-norm fuzzy logics, as well as their expansions by modal and other operators, have been studied thoroughly. Important examples of t-norm fuzzy logics are monoidal t-norm logic of all left-continuous t-norms, basic logic of all continuous t-norms, product fuzzy logic of the product t-norm, or the nilpotent minimum logic of the nilpotent minimum t-norm. Some independently motivated logics belong among t-norm fuzzy logics as well, like {\L}ukasiewicz logic (which is the logic of the {\L}ukasiewicz t-norm) and G\"odel-Dummett logic (which is the logic of the minimum t-norm).

Recent results \cite{bounded,boundedeq,boundedimpl,aggr,ijcci,iwobi} show that in the field of continuous logic, nilpotent logical systems are the most suitable for neural computation, mainly because of their bounded generator functions. Moreover, among other preferable properties, the fulfillment of the law of contradiction and the excluded middle, and the coincidence of the residual and the S-implication \cite{Dubois, Trillasimpl} also make the application of nilpotent operators in logical systems promising. 
In \cite{bounded,boundedeq,boundedimpl,aggr,ijcci,iwobi} a rich asset of operators were examined thoroughly:
in \cite{bounded}, negations, conjunctions and disjunctions, in \cite{boundedimpl} implications, and in \cite{boundedeq} equivalence operators. In \cite{aggr}, a parametric form of a general operator $o_{\nu}$ was given by using a shifting transformation of the generator function. 
Varying the parameters, nilpotent conjunctive, disjunctive, aggregative (where a high input can compensate for a lower one) and negation operators can all be obtained.
Moreover, as it was shown in \cite{ijcci}, membership functions, which play a substantial role in the overall performance of fuzzy representation, can also be defined by means of a generator function. 

In this study, we introduce a nilpotent neural model, where nilpotent logical operators and multicriteria decision tools are implemented in the hidden layers of neural networks (see Figure \ref{fig:nnm}). Only the weights of the first layer (parameters of hyperplanes separating the decision space) are to be learned, and the architecture needs to be designed. In a more sophisticated version, left for future work, the type of the operators in the hidden layers can also be learned by the network, or e.g. by a genetic algorithm. Moreover, in \cite{ijcci} the authors showed that the most important logical operators can be expressed as a composition of a parametric unary operator and the arithmetic mean. This means that the neural network only needs to learn the parameters of the first layer and (if not initially given) the parameters of these unary operators in the hidden layers.

In the nilpotent neural model, the activation functions in the first layer are membership functions representing truth values of inequalities, normalizing the inputs. At the same time, the activation functions in the the hidden layers model the cutting function (or to avoid the vanishing gradient problem, its differentiable approximation, the so-called squashing function) in the nilpotent logical operators. The theoretical background offers a straightforward choice of activation functions: the squashing function, which is an approximation of the rectifier. The fact that the squashing function, in contrast to the rectifier, is bounded from above, makes the continuous logical concept applicable.

The article is organized as follows. After summarizing the most important related work in Section \ref{relwork}, we revisit the relevant preliminaries concerning nilpotent logical systems in Section \ref{nilpot}. The nilpotent neural concept is described in Section \ref{sec3}. In Section \ref{PE}, the model is illustrated with some extended tensorflow playground examples.	Finally, the main results are summarized in Section \ref{concl}.

\section{Related Work} \label{relwork}

Combinations of neural networks and logic rules have been considered in different contexts. 
Neuro-fuzzy systems \cite{neurofuzzy} were examined thoroughly in the literature. These hybrid intelligent systems synergize the human-like reasoning style of fuzzy systems with the learning structure of neural networks through the use of fuzzy sets and a linguistic model consisting of a set of IF-THEN fuzzy rules. These models were the first attempts to combine continuous logical elements and neural computation.

KBANN \cite{Towell}, Neural-symbolic systems \cite{Garcez}, such as CILP++ \cite{Franca}, constructed network architectures from given rules to perform knowledge acquisition. 

Kulkarni et al. \cite{Kulkarni} used a specialized training procedure to obtain an interpretable neural layer of an image network. 

In \cite{harness}, Hu et al. proposed a general framework capable of enhancing various types of neural networks (e.g., CNNs and RNNs) with declarative first-order logic rules. Specifically, they developed an iterative distillation method that transfers the structured information of logic rules into the weights of neural networks. With a few highly intuitive rules, they obtained substantial improvements and achieved state-of-the-art or comparable results to previous best-performing systems.

In \cite{xu}, Xu et al. developed a novel methodology for using symbolic knowledge in deep learning by deriving a semantic loss function that bridges between neural output vectors and logical constraints. This loss function captures how close the neural network is to satisfying the constraints on its output. 

In \cite{dl2}, Fischer et al. presented DL2, a system for training and querying neural networks with logical constraints. Using DL2, one can declaratively specify domain knowledge constraints to be enforced during training, as well as pose queries on the model to find inputs that satisfy a set of constraints. DL2 works
by translating logical constraints into a loss function with desirable mathematical properties. The loss is then minimized with standard gradient-based methods. 

All of these promising approaches point towards the desirable mathematical framework that nilpotent logical systems can offer. Our general aspiration here is to provide a general mathematical framework in order to benefit from a tight integration of machine learning and continuous logical methods.

\begin{figure*}[t]
	\includegraphics[width=0.9\textwidth]{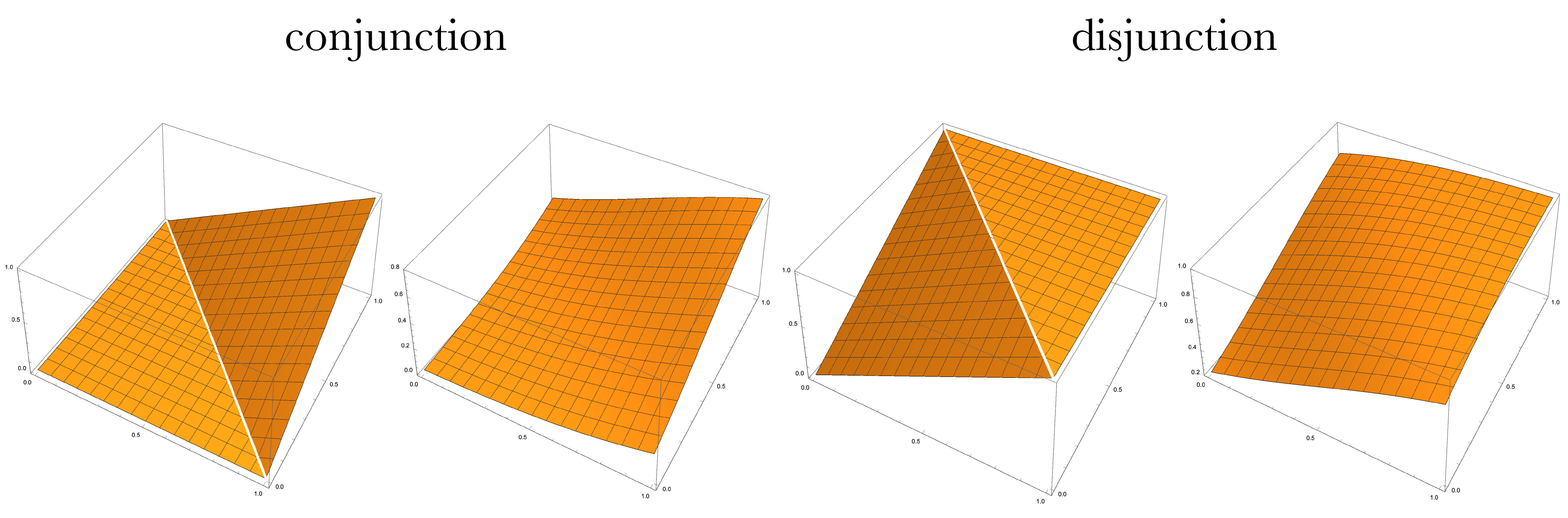}
	\caption{Nilpotent conjunction and disjunction followed by their approximations using the squashing function}
	\label{fig:conj}
\end{figure*}

\section{Nilpotent Logical Systems and Multicriteria Decision Tools} \label{nilpot}
In this Section, we show why a specific logical system, the nilpotent logical system is well-suited to the neural environment.
First, we provide some basic preliminaries.

The most important operators in classical logic are the conjunction, the disjunction and the negation operator. These three basic operators together form a so-called connective system. When extending classical logic to continuous logic, compatibility and consistency are crucial.  The negation should also be involutive; i.e. $n(n(x))=x,$ for $\forall x \in [0,1].$ Involutive negations are called strong negations.
\begin{thm}
	The triple $(c,d,n),$ where $c$ is a t-norm, $d$ is a t-conorm and $n$ is a strong negation, is called a connective system.
\end{thm}

As mentioned in the Introduction, numerous continuous logical systems have been introduced and studied in the literature. In this study, we will show how nilpotent logical systems relate to neural networks.
\begin{thm}\cite{bounded}
	A connective system is nilpotent, if the conjunction $c$ is a nilpotent t-norm, and the disjunction $d$ is a nilpotent t-conorm.
\end{thm}

In the nilpotent case, the generator functions of the disjunction and the conjunction (denoted by $t(x)$ and $s(x)$ respectively) are bounded functions, being determined up to a multiplicative constant. This means that they can be normalized the following way:
\begin{equation}
f_{c}(x):=\frac{t(x)}{t(0)},\quad \quad f_{d}(x):=\frac{s(x)}{s(1)}.
\end{equation}

	Note that the normalized generator functions are now uniquely defined. 

Next, we recall the definition of the cutting function, to simplify the notations used. The differentiable approximation of this cutting function, the squashing function $S(x)$ introduced and examined in \cite{Gera}, will be a ReLu-like bounded activation function in our model. In \cite{aggr}, the authors showed that all the nilpotent operators can be described by using one generator function $f(x)$ and the cutting function.

\begin{thm}
	Let us define the cutting operation $[\phantom{x}]$ by
	\begin{center}
		\[[x]=\left\{
		\begin{array}
		[c]{ccc}%
		0 & if & x<0\\
		x & if & 0\leq x\leq 1\\
		1 & if & 1<x\\
		\end{array}
		\right. \]
	\end{center}
\end{thm}
\begin{rem}
	Note that the cutting function has the same values as ReLu (rectified linear unit) for $x\leq 1,$ but it remains bounded for $\forall x\in \mathbb{R}$.
\end{rem}
\begin{pro} \label{cutth}
	With the help of the cutting operator, we can write the conjunction and disjunction in the following form, where $f_c$ and $f_d$ are decreasing and increasing normalized generator functions respectively. 
	\begin{equation} \label {[]} c(x,y)=f_{c}^{-1}[f_{c}(x)+f_{c}(y)],\end{equation}
	\begin{equation} \label{[]2} d(x,y)=f_{d}^{-1}[f_{d}(x)+f_{d}(y)].\end{equation}\end{pro}
\begin{rem}
	For the natural negations to coincide, as shown in \cite{bounded}, $f_{c}(x)+f_{d}(x)=1$ must hold for $\forall x \in [0,1],$ which means that only one generator function, e.g. $f_d(x)$ is needed to describe the operators. Henceforth,  $f_d$ is represented by $f(x)$.
\end{rem}
\begin{rem}
Note that the $\min$ and $\max$ operators (often used as conjunction and disjunction in applications) can also be expressed by $[\phantom{x}]$ in the following way:
	\begin{equation}
	\min(x,y)=\left[x+[y-x+1]-1\right],\label{min}
	\end{equation}
	\begin{equation}
	\max(x,y)=\left[x+[y-x]\right],\label{max}
	\end{equation}
	where $x,y \in [0,1].$
\end{rem}
The associativity of t-norms and t-conorms permits us to consider their extensions to the multivariable case.
In \cite{aggr}, the authors examined a general parametric operator $o_{\nu}(\underline{x})$ of nilpotent systems.
\begin{table*}[t]
	\caption{The most important two-variable operators $o_{\underline{w}} (\underline{x})$}\label{abc}
	\begin{center}
		\begin{tabular}{ | l | l | l | l | l | l | l |}
			\hline
			& $w_1$ & $w_2$ & $C$ & $o_{\underline{w}}(x,y)$&for $f(x)=x$ &Notation  \\ \hline \hline
			\multicolumn{7}{|l|}{LOGICAL OPERATORS}\\	\hline
			disjunction& $1$ & $1$ & $0$  & $f^{-1}[f(x)+f(y)]$  &$[x+y]$ & $d(x,y)$ \\ \hline
			conjunction& $1$ & $1$ & $-1$  & $f^{-1}[f(x)+f(y)-1]$  &$[x+y-1]$& $c(x,y)$ \\ \hline
			implication& $-1$ & $1$ & $1$  & $f^{-1}[f(y)-f(x)+1]$  &$[y-x+1]$& $i(x,y)$ \\ \hline
			\multicolumn{7}{|l|}{MULTICRITERIA DECISION TOOLS}\\	\hline
			arithmetic mean& $0.5$ & $0.5$ & $0$  & $f^{-1}\left[\frac{f(x)+f(y)}{2}\right]$  &$\frac{x+y}{2}$ & $m(x,y)$ \\ \hline
			preference & $-0.5$ & $0.5$ & $0.5$  & $f^{-1}\left[\frac{f(y)-f(x)+1}{2}\right]$  & $\frac{y-x+1}{2}$ &$p(x,y)$ \\ \hline
			aggregative operator & $1$ & $1$ & $-0.5$  & $f^{-1}\left[f(x)+f(y)-\frac{1}{2}\right]$  & $\left[x+y-\frac{1}{2}\right]$ & $a(x,y)$ \\ \hline
		\end{tabular}
	\end{center}
\end{table*} 

\begin{thm}Let $f:[0,1]\rightarrow [0,1]$ be an increasing bijection, $\nu\in[0,1]$, and $\underline{x}=(x_1,\dots, x_n),$ where $x_i\in[0,1]$ and let us define the general operator by
	\begin{equation} \label{o}
	\begin{split}
	o_{\nu}(\underline{x})=f^{-1}\left[\sum\limits_{i=1}^n \left(f(x_i)-f(\nu)\right)+f(\nu)\right]=
	\\
	=f^{-1}\left[\sum\limits_{i=1}^n f(x_i)-(n-1)f(\nu)\right].
	\end{split}
	\end{equation}
\end{thm}

\begin{rem} \label{cd} Note that the general operator for $\nu=1$ is conjunctive, for $\nu=0$ it is disjunctive and for $\nu=\nu^*=f^{-1}\left(\frac{1}{2}\right)$ it is self-dual.
\end{rem}

\begin{table*}[h!]
	\caption{The most important unary operators $o_{\alpha, \gamma}(x)$}\label{ac}
	\begin{center}
		\begin{tabular}{ | l |  l | l | l | l |}
			\hline
			& $\gamma$ & $o_{\alpha, \gamma}(x)$& for $f(x)=x$&Notation  \\ \hline \hline
			possibility& $0$  & $f^{-1}[\alpha f(x)]$  & $[\alpha x]$  &$\tau_{P}(x)$ \\ \hline
			necessity & $1-\alpha$  & $f^{-1}[\alpha f(x)-(\alpha-1)]$  &  $[\alpha x-(\alpha-1)]$&$\tau_N(x)$ \\ \hline
			sharpness & $\frac{\alpha-1}{2}$  & $f^{-1}[\alpha f(x)-\frac{(\alpha-1)}{2}]$  &$[\alpha x-\frac{(\alpha-1)}{2}]$ &$\tau_S(x)$ \\ \hline
		\end{tabular}
	\end{center}
\end{table*} 

On the basis of Remark \ref{cd}, the conjunction, the disjunction and the aggregative operator can be defined in the following way.

\begin{thm}Let $f:[0,1]\rightarrow [0,1]$ be an increasing bijection, $\underline{x}=(x_1,\dots, x_n),$ where $x_i\in[0,1]$. Let us define the conjunction, the disjunction and the aggregative operator by
	\begin{equation}c(\underline{x}):=o_1(\underline{x})=f^{-1}\left[\sum\limits_{i=1}^n f(x_i)-(n-1)\right],\label{7}\end{equation}
	\begin{equation}d(\underline{x}):=o_0(\underline{x})=f^{-1}\left[\sum\limits_{i=1}^n f(x_i)\right],\label{8}\end{equation}
	\begin{equation}a(\underline{x}):=o_{\nu^*}(\underline{x})=f^{-1}\left[\sum\limits_{i=1}^n f(x_i)-\frac{(n-1)}{2}\right],\label{9}\end{equation}
	respectively, where $\nu^*=f^{-1}\left(\frac{1}{2}\right)$.
\end{thm}

	A conjunction, a disjunction and an aggregative operator differ only in one parameter of the general operator in \eqref{o}. The parameter $\nu$ has the semantic meaning of the level of expectation: maximal for the conjunction, neutral for the aggregation and minimal for the disjunction. 
Next, let us recall the weighted form of the general operator:
\begin{thm}Let $\underline{w}\in \mathbb{R}^n,
	 f:[0,1]\rightarrow [0,1]$ an increasing bijection with $\nu\in[0,1], \underline{x}=(x_1,\dots, x_n),$ where $x_i\in[0,1].$ The weighted general operator is defined by
	\begin{equation}\label{a}
	o_{\nu,\underline{w}}(\underline{x}):=f^{-1}\left[\sum\limits_{i=1}^n w_i (f(x_i)-f(\nu))+f(\nu)\right].
	\end{equation}
\end{thm}

Note that if the weight vector is normalized; i.e. for $\sum_{i=1}^{n} w_i=1,$ 
\begin{equation}
	o_{\nu,\underline{w}}(\underline{x})=f^{-1} \left(\sum_{i=1}^{n} w_i f(x_i)\right).
\end{equation}
For future application, we introduce a threshold-based operator in the following way.
\begin{thm} Let $\underline{w}\in \mathbb{R}^n, \underline{x}=(x_1,...x_n)\in [0,1]^n$, $\underline{\nu}=(\nu_1,...\nu_n) \in [0,1]^n$and let $f:[0,1]\rightarrow[0,1]$ be a strictly increasing bijection. Let us define the threshold-based nilpotent operator by
	\begin{equation*}
o_{\underline{\nu},\underline{w}}(\underline{x})=
	f^{-1} \left[\sum_{i=1}^{n} w_i \left(f(x_i)-f(\nu_i\right))+f(\nu)\right]=
	\end{equation*}
\begin{equation}	=f^{-1} \left[\sum_{i=1}^{n} w_i f(x_i)+C\right],\label{4}
	\end{equation}
\end{thm} 
where 
\begin{equation}
C = f(\nu)-\sum_{i=1}^{n} w_i f(\nu_i).
\end{equation}

	Note that for $f(x)=x$, \eqref{4} gives the functions modeled by perceptrons in neural networks:
\begin{equation}
 \left[\sum_{i=1}^{n} w_i x_i+C\right].\label{5}
\end{equation}	
	Based on Equations \eqref{7} to \eqref{9}, it is easy to see  that the conjunction, the disjunction and also the aggregative operator can be expressed in this form. The most commonly used operators for $n=2$ and for special values of $w_i$ and $ C$, also for $f(x)=x$, are listed in Table \ref{abc}.

Now let us focus on the unary (1-variable) case, examined in \cite{ijcci}, which also plays an important role in the nilpotent neural model. The unary operators are mainly used to construct modifiers and membership functions by using a generator function. The membership functions can be interpreted as modeling an inequality \cite{memeva}. Note that non-symmetrical membership functions can also be constructed by connecting two unary operators with a conjunction \cite{iwobi,ijcci}.

\begin{thm}
	Let $x \in [0,1]$, $\alpha, \gamma \in \mathbb{R}$ and let $f:[0,1]\rightarrow[0,1]$, a strictly increasing bijection. Then
	\begin{equation}o_{\alpha, \gamma}(x):=f^{-1}[\alpha f(x)+\gamma].\label{unary}\end{equation}
\end{thm}

\begin{rem}
	Note that as shown in \cite{ijcci}, Equation \eqref{unary} composed by the (weighted) arithmetic mean operator as an inner function, yields to Equation \eqref{4}.
\end{rem}	 The most important unary operators for special $\gamma$ values are listed in Table \ref{ac}.

Our attention can now be turned to the cutting function. The main drawback of the cutting function in the nilpotent operator family is the lack of differentiability, which would be necessary for numerous practical applications. Although most fuzzy applications (e.g. embedded fuzzy control) use piecewise linear membership functions owing to their easy handling, there are areas where the parameters are learned by a gradient-based optimization method. In this case, the lack of continuous derivatives makes the application impossible. For example, the membership functions have to be differentiable for each input in order to fine-tune a fuzzy control system by a simple gradient-based technique.
This problem could be easily solved by using the so-called squashing function, which provides a solution to the above-mentioned problem by a continuously differentiable approximation of the cutting function.

\begin{figure}
	\begin{center}
		\includegraphics[width=0.35\textwidth]{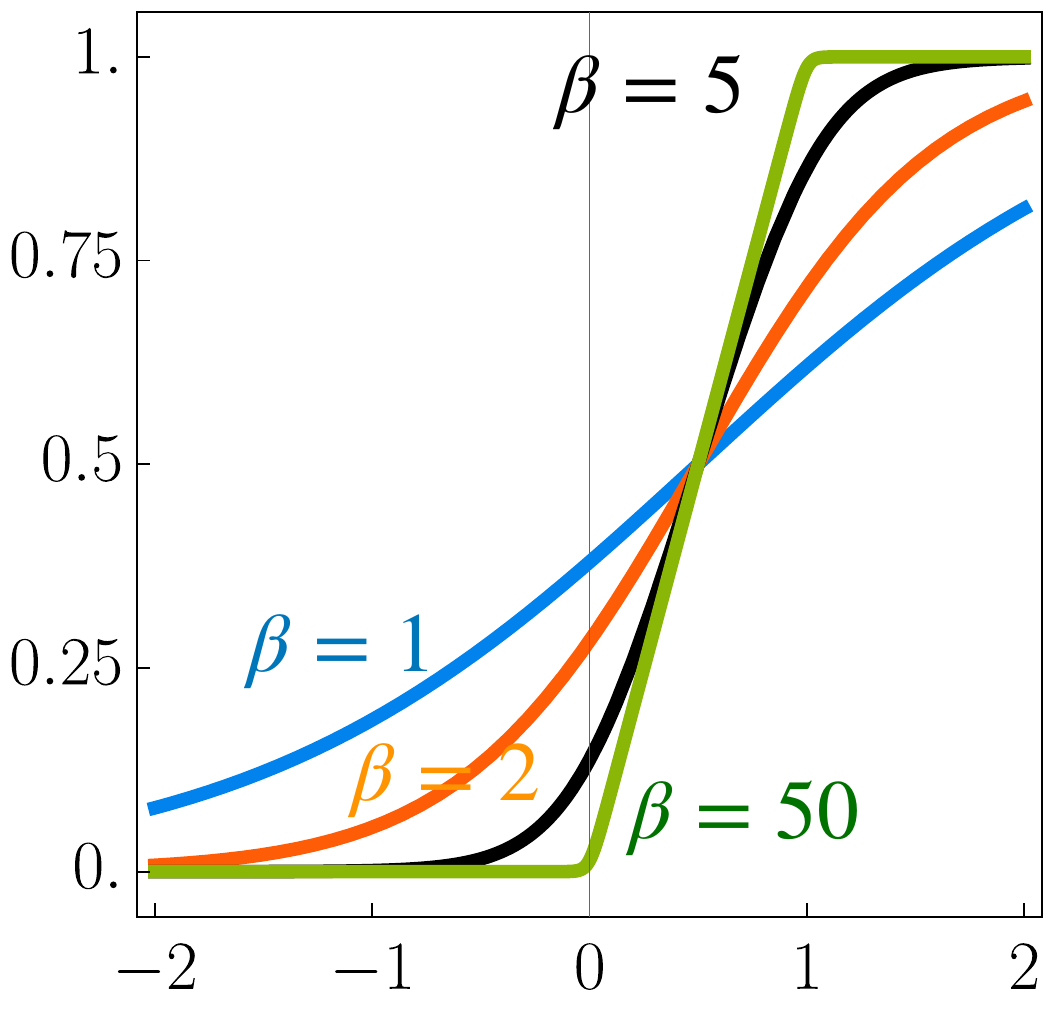}
	\end{center}
	\caption{Squashing functions for $a=0.5$, $\lambda=1,$ for different $\beta$ values ($\beta_1=1,$ $\beta_2=2, $$\beta_3=5, $ and $\beta_4=50$) }\label{fig:squash}
\end{figure}

The squashing function given in Definition \ref{squashdef} is a continuously differentiable approximation of the generalized cutting function by means of sigmoid functions (see Figure \ref{fig:squash}).  

\begin{thm}
	The squashing function \cite{Gera, ijcci} is defined as 
	\begin{equation*}
	S^{(\beta)}_{a,\lambda}(x) = \frac1{\lambda\beta}\ln\frac{1+e^{\beta\left(x-(a-\lambda/2)\right)}}{1+e^{\beta\left(x-(a+\lambda/2)\right)}} = \frac1{\lambda\beta}\ln\frac{\sigma_{a+\lambda/2}^{(-\beta)}(x)}{\sigma_{a-\lambda/2}^{(-\beta)}(x)}.
	\end{equation*}
	where $x,a,\lambda,\beta\in\mathbb{R}$ and $\sigma_d^{(\beta)}(x)$ denotes the logistic function:
	\begin{align}
	\sigma_d^{(\beta)}(x) = \frac1{1+e^{-\beta\cdot (x-d)}}.
	\end{align}
	\label{squashdef}
\end{thm}
By increasing the value of $\beta$, the squashing function approaches the generalized cutting function. In other words, $\beta$ shows the accuracy of the approximation, while the parameters $a$ and $\lambda$ determine the center and width.
The error of the approximation can be upper bounded by $c/\beta$, which means that by increasing the parameter $\beta$, the error decreases by the same order of magnitude.
The derivatives of the squashing function are easy to calculate and can be expressed by sigmoid functions and itself:

\begin{align}
\frac{\partial S^{(\beta)}_{a,\lambda}(x)}{\partial x} &= \frac1\lambda\left(\sigma_{a-\lambda/2}^{(\beta)}(x)-\sigma_{a+\lambda/2}^{(\beta)}(x)\right) \\
\frac{\partial S^{(\beta)}_{a,\lambda}(x)}{\partial a} &= \frac1\lambda\left(\sigma_{a+\lambda/2}^{(\beta)}(x)-\sigma_{a-\lambda/2}^{(\beta)}(x)\right) \\
\frac{\partial S^{(\beta)}_{a,\lambda}(x)}{\partial \lambda} &=
-\frac1\lambda{}S^{(\beta)}_{a,\lambda}(x)+\frac1{2\lambda}\left(\sigma_{a+\lambda/2}^{(\beta)}(x)+\sigma_{a-\lambda/2}^{(\beta)}(x)\right)
\end{align}

The squashing function defined above is an approximation of the rectifier (rectified linear unit, ReLU) for $x\leq1$, with the benefit of having an upper bound. Being bounded from above makes the use of continuous logic possible. Also note the significant difference between the properties of the squashing function and the sigmoid. Using sigmoids, nilpotent logic can never be modeled. The fact that on the other hand, ReLu can approximate the cutting function, may offer an interpretation to its effectiveness and success.
The fact that the squashing function is differentiable and its derivatives can be expressed by sigmoids improves efficiency in applications.
An illustration of the nilpotent conjunction and disjunction operators with their soft approximations using the squashing function are shown in Figure \ref{fig:conj}. 
Note that not only logical operators, but also multicriteria decision tools, like the preference operator can be described similarly. This means that our model offers a unified framework, in which logic and multicriteria decision tools cooperate and supplement each other.

\section{Nilpotent Logic-based Interpretation of Neural Networks}\label{sec3}
The results on nilpotent logical systems discussed in Section \ref{nilpot} offer a new approach to designing neural networks using continuous logic, since membership functions (representing the truth value of an ineqaulity), and also nilpotent logical operators can be modeled by perceptrons. Whether for image classification or for multicriteria decision support, structured logical rules can contribute to the performance of a deep neural network. Given that the network has to find a region in the decision space or in an image, after designing the architecture appropriately, the network only has to find the parameters of the boundary. Here, we propose creating basic building blocks by applying the nilpotent logical concept in the perceptron model and also in the neural architecture.

\begin{figure}
	\center\includegraphics[width=0.4\textwidth]{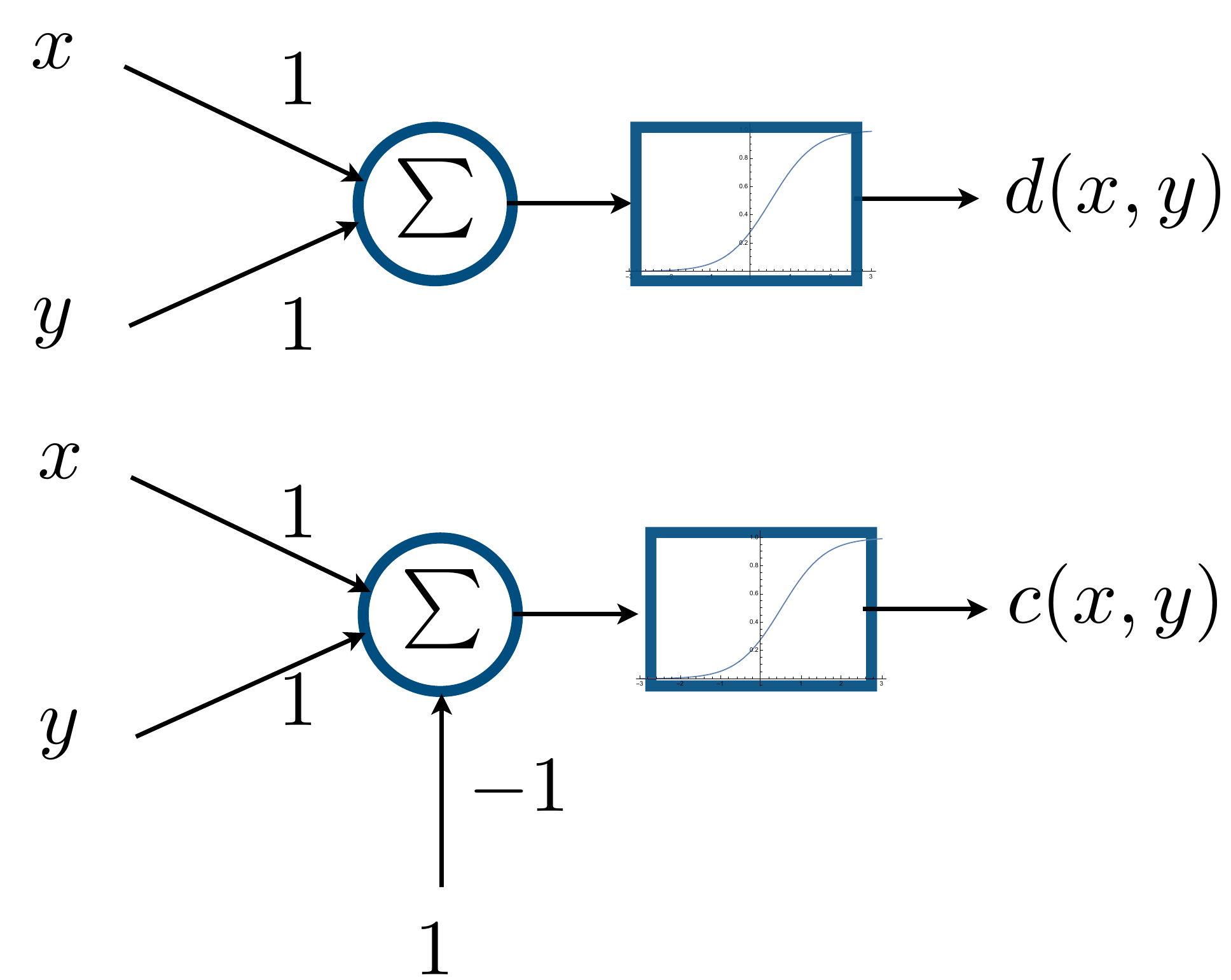}
	\caption{Perceptron model of the conjunction and the disjunction}
	\label{fig:cdfig}
\end{figure}

Boolean units and multilayer perceptrons have a long history. Logical gates (such as the AND, NOT and OR gates) are the basis of any modern day computer.
It is well known that any Boolean function can be composed using a multi‐layer perceptron. 
As examples, the 
conjunction and the disjunction are illustrated in Figure 
\ref{fig:cdfig}. Note that for the XOR gate, an additional hidden layer is also required. It can be shown that 
a network of linear classifiers that fires if the input is in a given area with arbitrary complex decision boundaries can be constructed with only one hidden layer and a single output. 
This means that if a neural network learns to separate different regions in the $n$-dimensional space having $n$ input values, each node in the first layer can separate the space into two half-spaces by drawing one hyperplane, while the nodes in the hidden layers can combine them using logical operators.

\begin{figure}
	\begin{center}
		\includegraphics[width=0.48\textwidth]{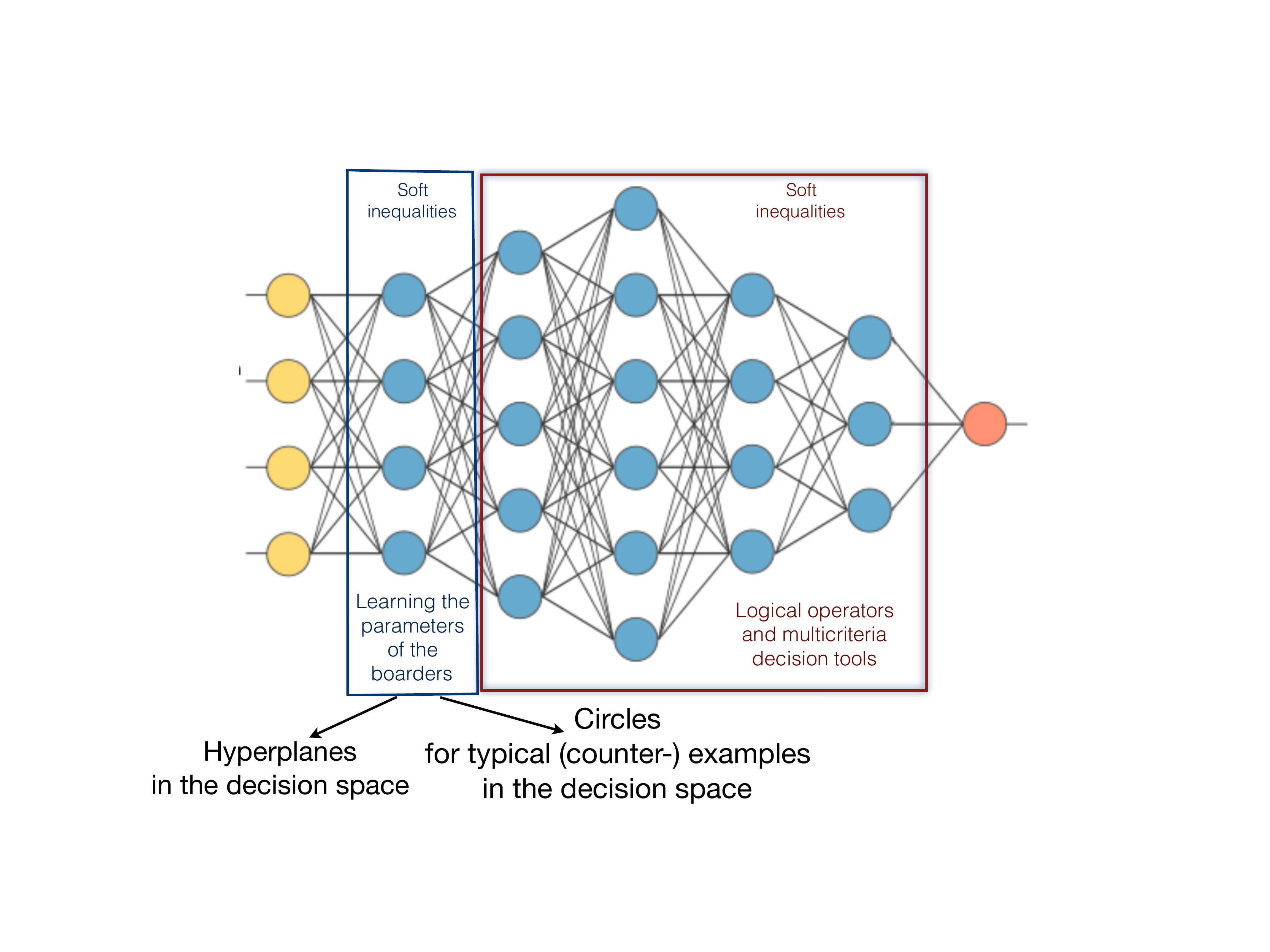}
		\caption{Nilpotent neural model}	\label{fig:nnm}
	\end{center}
\end{figure}

\begin{figure}
	\includegraphics[width=0.48\textwidth]{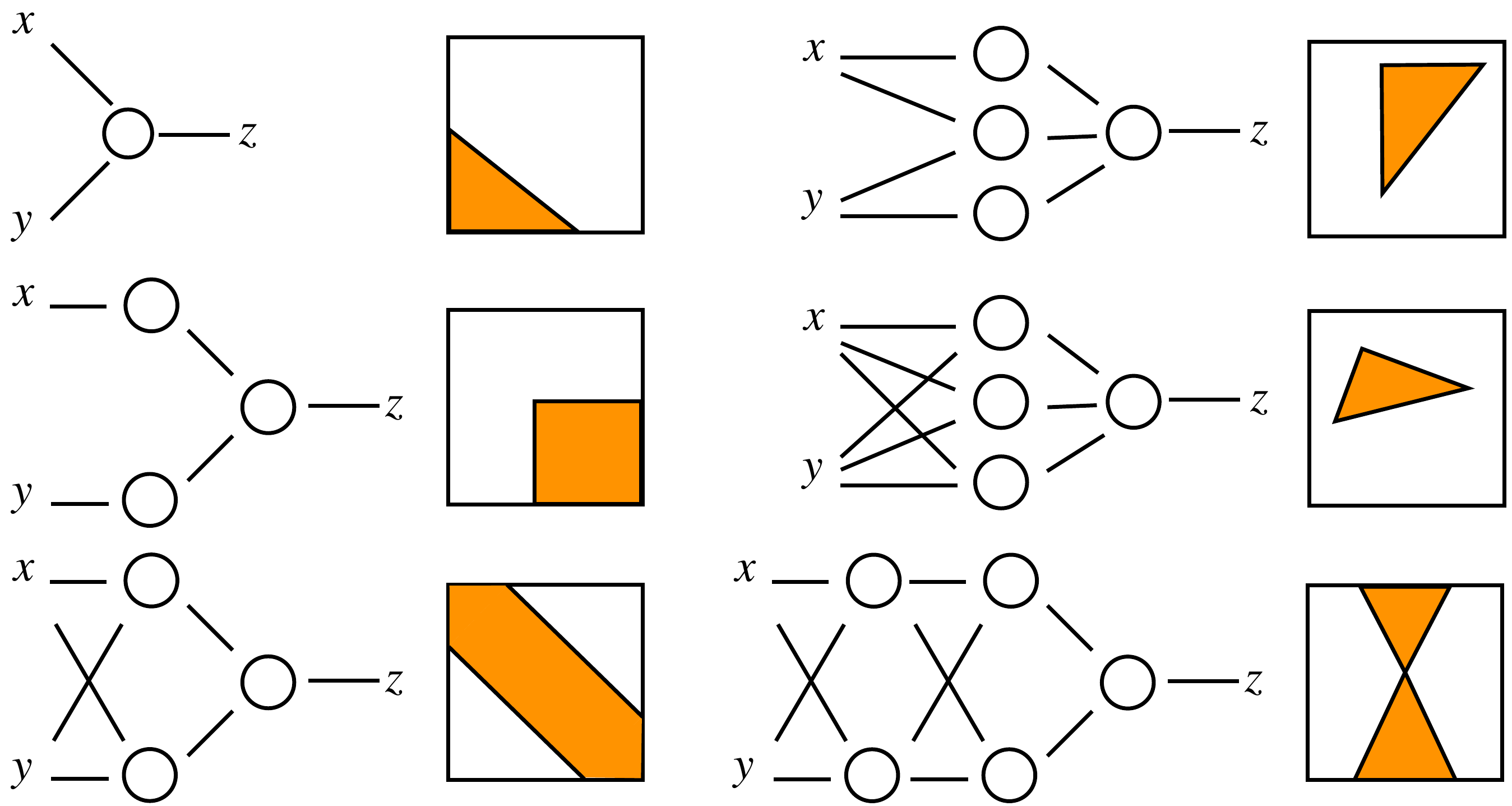}
	\caption{Basic types of neural networks with two input values using logical operators in the hidden layer used to find different regions of the plane}
	\label{fig:mit}
\end{figure}

In Figure \ref{fig:mit}, some basic types of neural networks are shown with two input values, finding different regions of the plane. Generally speaking, each node in the neural net represents one threshold and therefore it can draw one line in the picture. The line may be diagonal if the nodes receives both of the inputs $i_1$ and $i_2$. The line has to be horizontal or vertical if the node only receives one of the inputs. The deeper hidden levels are responsible for the logical operations.

From several perspectives, as mentioned in the Introduction, a continuous logical framework can provide a more sophisticated and effective approach to this problem than a boolean can. 

\begin{figure}
	\begin{center}
		\includegraphics[width=0.45\textwidth]{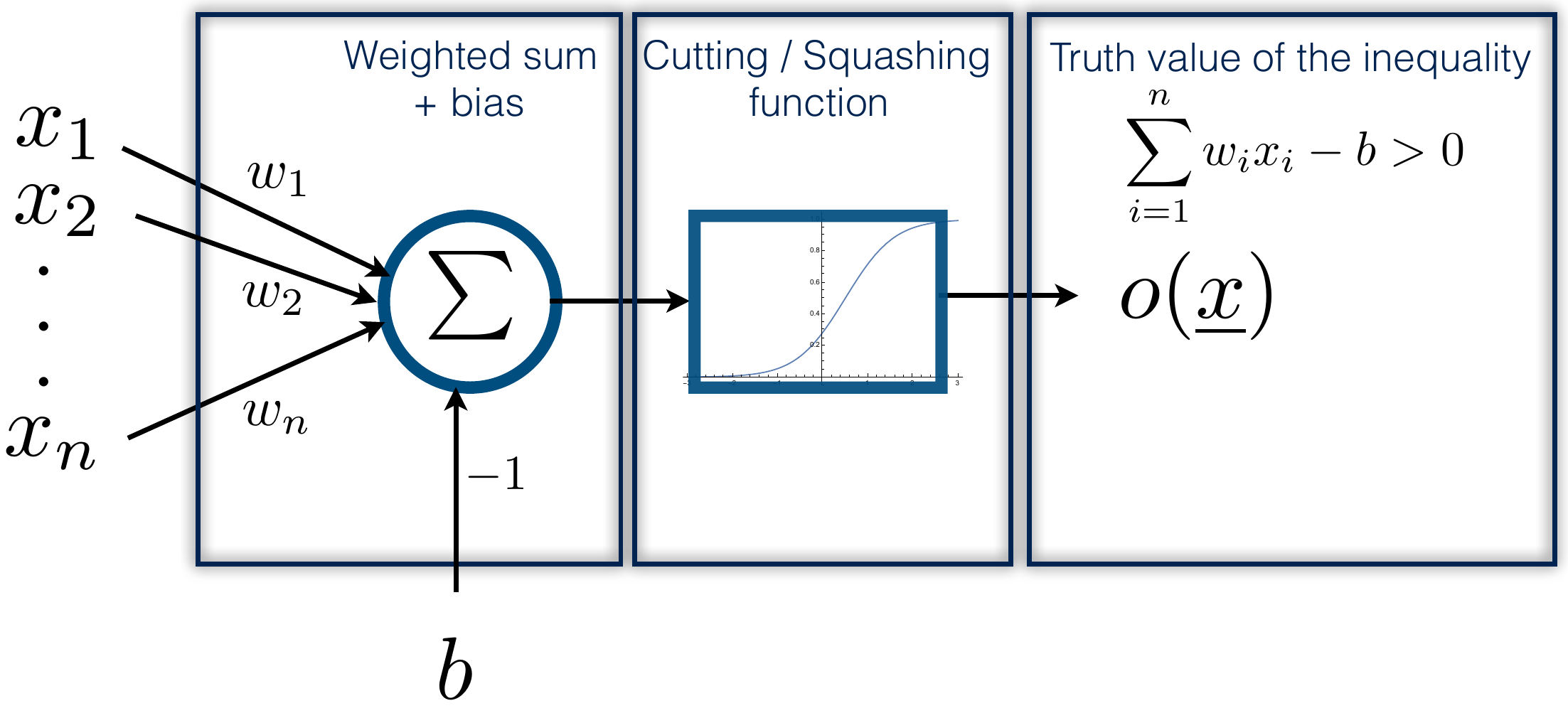}
		\caption{Nilpotent perceptron model}	\label{fig:nnm_perc}
	\end{center}
\end{figure}

Among continuous logical systems, the nilpotent logical framework described above is well-suited for the neural concept architecture, when it comes to implementing logical rules. For the sake of simplicity, henceforth we assume that for the generator function $f(x)=x$ holds and we design the neural network architecture in the following way. 
In the first layer, the perceptrons model membership functions as truth values of inequalities, such as \begin{equation}\sum_{i=1}^{n} w_{i}x_{i}-b>0;\label{ineq}\end{equation} representing a half space bounded by a hyperplane in the decision space (see Figure \ref{fig:nnm_perc}). Here, the weights $w_i$  and the bias $b$ are to be learned. The truth value of this inequality can be modeled by \begin{equation}\left[\sum_{i=1}^{n} w_{i}x_{i}-b\right];\end{equation}
or to avoid the vanishing gradient problem, the cutting function can be approximated by the so-called squashing function, by the differentiable approximation of the cutting function 
\begin{equation}\label{s} S\left(\sum_{i=1}^{n} w_{i}x_{i}-b\right).\end{equation}
The parameters of the squashing function in \eqref{s} now have a  context-dependent semantic meaning. 

Since the nilpotent logical operators also represent inequalities and therefore have the same structure (compare with Equation \eqref{5}, and see also Table \ref{abc} and Figure \ref{fig:nnm_perc}), in the hidden layers, we can apply them on the clusters created in the first layer (see Figure \ref{fig:nnm}). Here,  the weights and biases characterize the type of the logical operator. As an illustration, the perceptron models of the conjunction and of the disjunction can be seen in Figure \ref{fig:cdfig}. This means that for a given logical operator, the weights and the bias can be frozen. The squashing function plays the role of the activation function in all of the layers.
The backpropagation algorithm needs to be adjusted: the error function is calculated based on all of the weights and biases (frozen and learnable), but the backpropagation leaves the frozen layers out.
Moreover, in this nilpotent model, the conjunction, the disjunction and the aggregation differ only in a translation parameter; i.e. the weights are equal for all of them and only the biases are different. This fact makes it possible for the network to learn the type of logical operators just by learning the bias. 

\begin{figure*}[!tb]
	\begin{center}$
		\begin{array}{cccc}
		\includegraphics[width=0.2\textwidth]{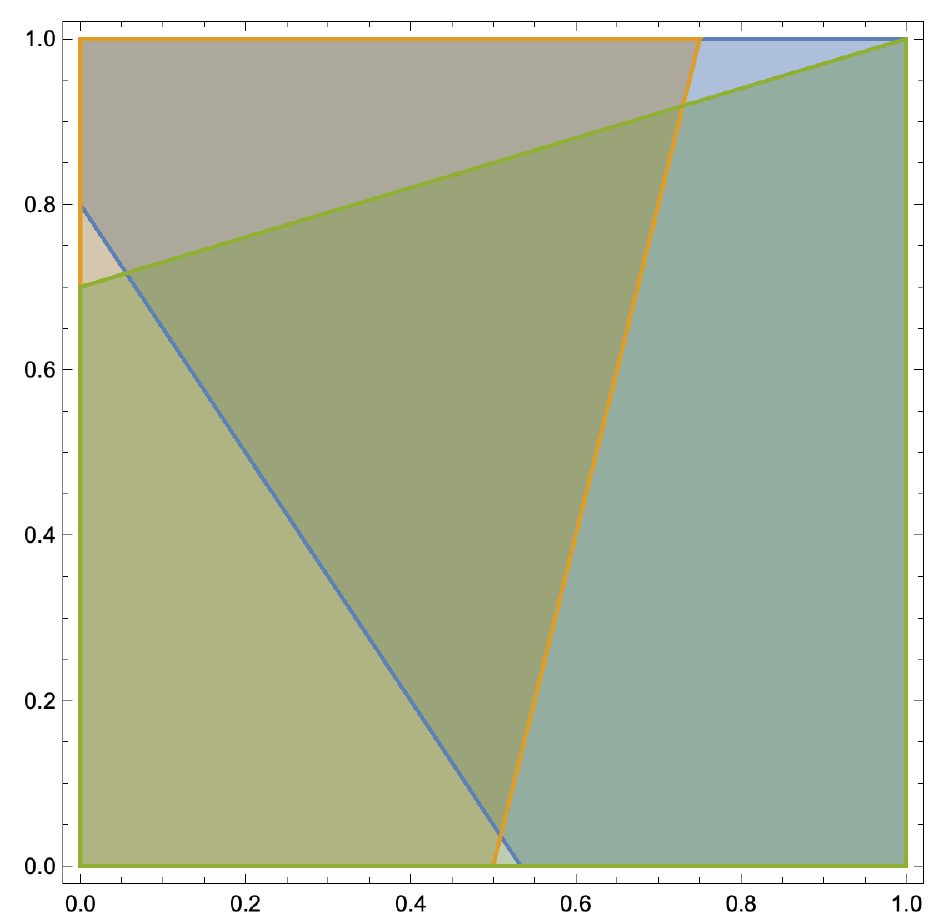}&
		\includegraphics[width=0.2\textwidth]{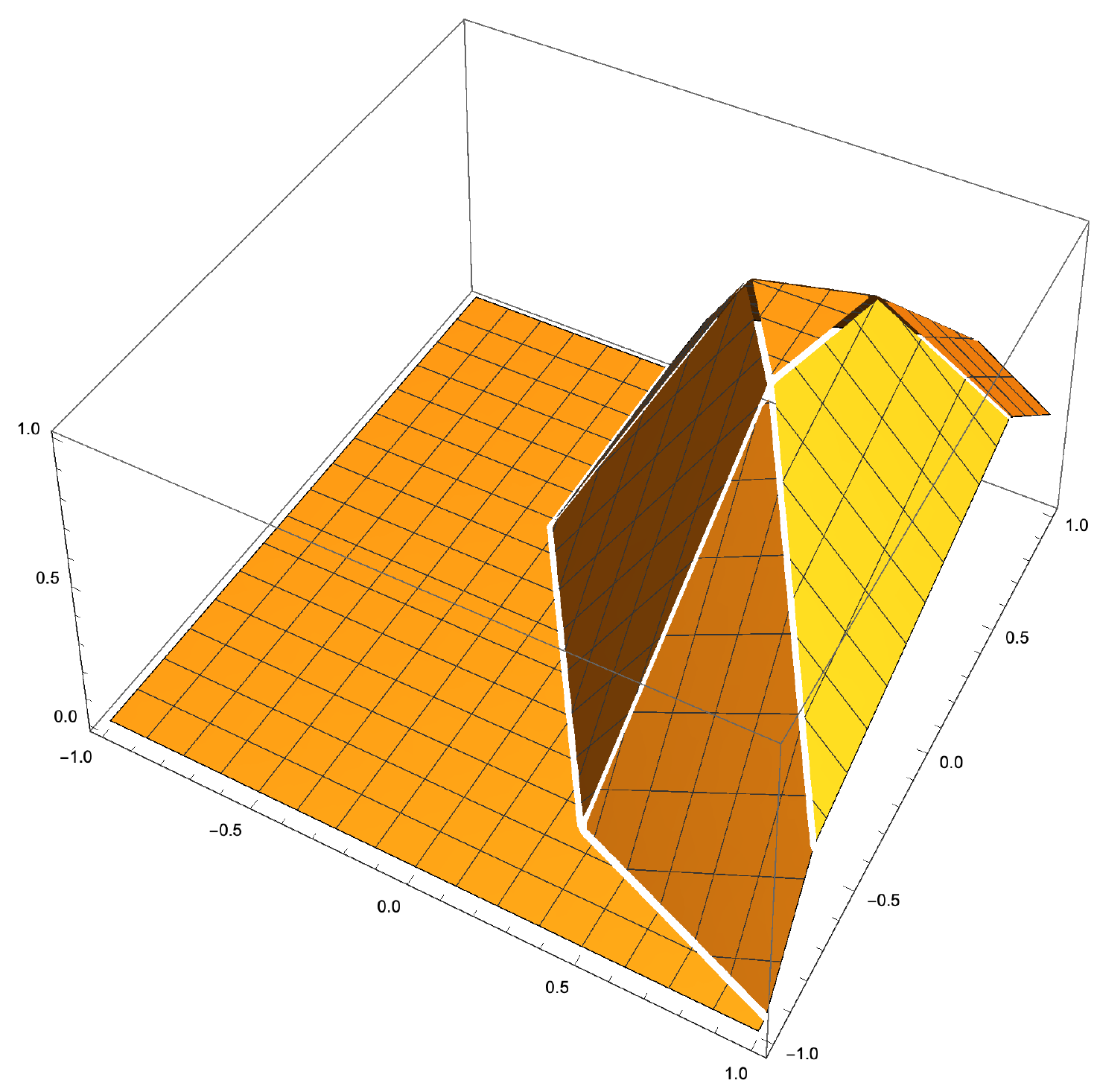}&	
		\includegraphics[width=0.2\textwidth]{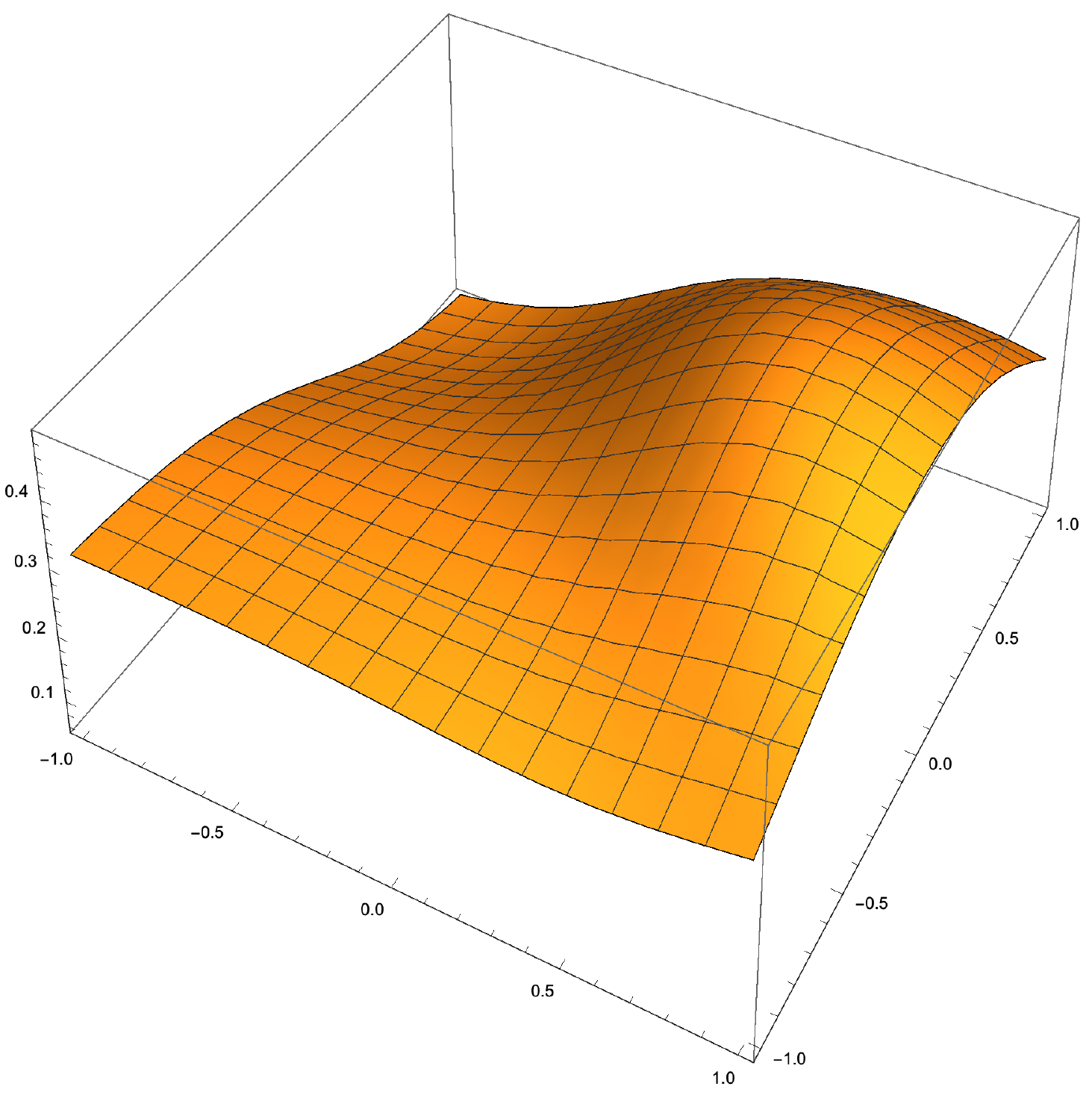}&
		\includegraphics[width=0.2\textwidth]{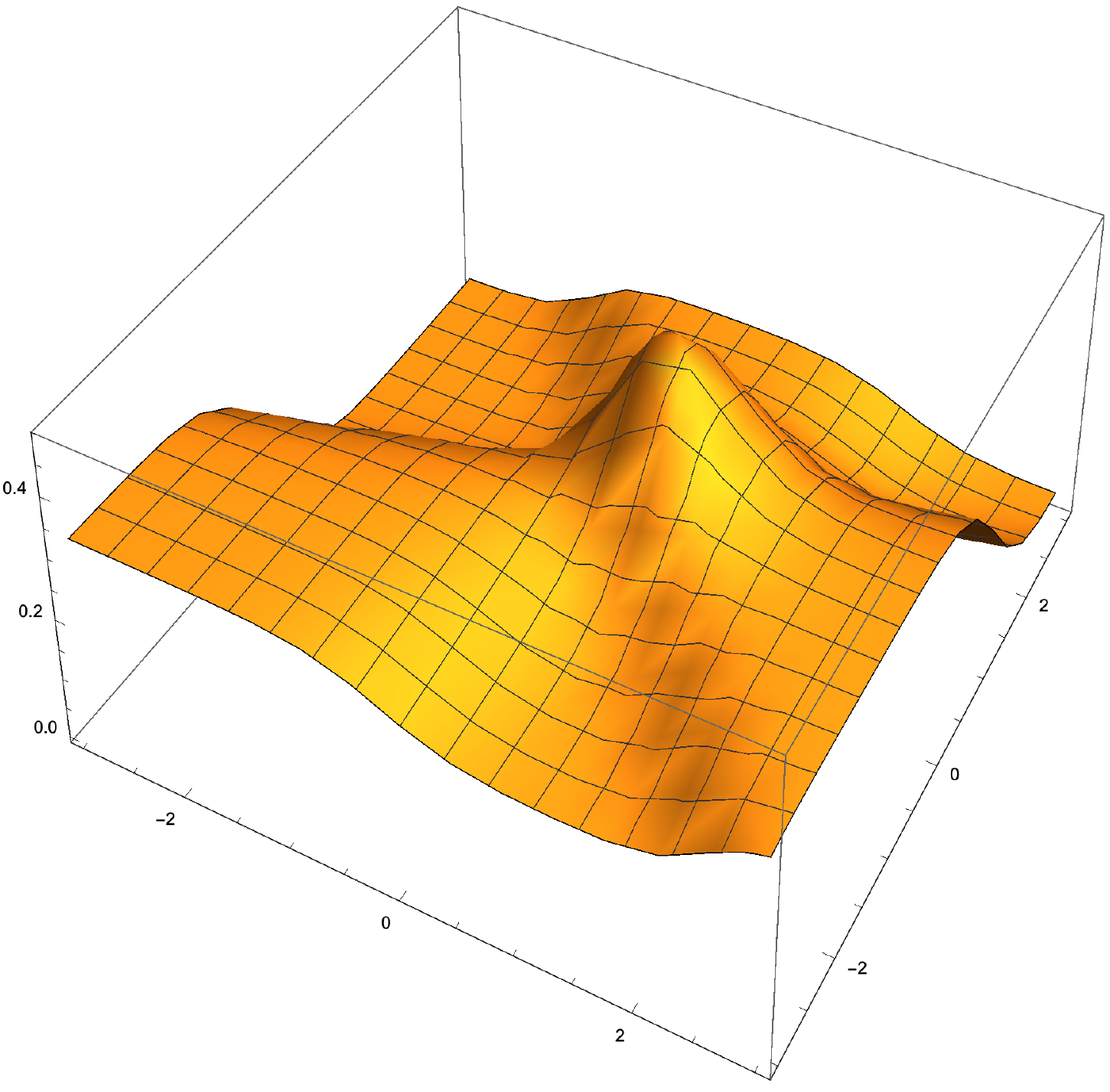}
		\end{array}$
	\end{center}
	\caption{Output values for a triangular domain using nilpotent logic and its continuous approximation for different parameter values}
	\label{fig:regions}		
\end{figure*}

\begin{figure}
	\begin{center}	\includegraphics[width=0.5\textwidth]{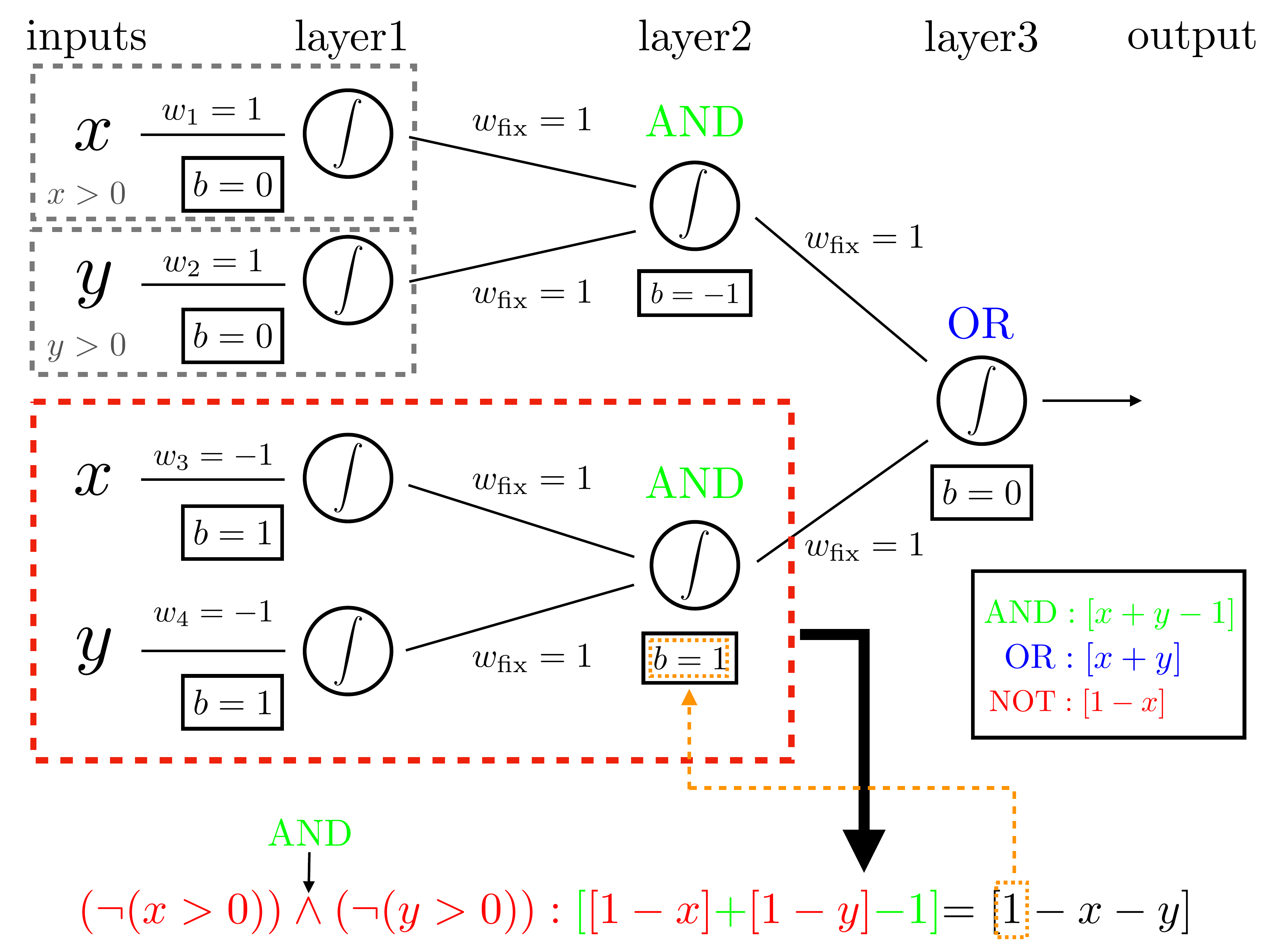}
		\caption{Nilpotent neural structure representing the expression  ($x>0) \mbox{ AND } (y>0) )\mbox{ OR } ( (x<0) \mbox{ AND } (y<0) $}\label{fig:structure}
	\end{center}
\end{figure}

To illustrate the model, two basic examples are given.
\begin{example}
	As an example, let us assume that a network needs to find positive examples which lie inside a triangular region. This means that we should design the network to conjunct three half planes, and to find the parameters of the boundary lines. The output values for a triangular domain using nilpotent logic and its continuous approximation are illustrated in Figure \ref{fig:regions}. 
\end{example}
\begin{example}\label{ex:E1}
	The flow chart and the model for the logical expression "$((x>0) \mbox{ AND } (y>0)) \mbox{ OR } ((x<0) \mbox{ AND } (y<0))$" can be seen in Figure \ref{fig:structure}. Here, $x<0$ is modeled by $\mbox{ NOT }(x>0)$. 
	Note that $\left[[x+y-1]+[(1-x)+(1-y)-1]\right]=$ $=\left[[x+y-1]+[-x-y+1]\right],$ therefore the bias for the negated inputs is $+1$. The weights and biases for the logical operators are listed in Table \ref{tab:xo}.
\end{example}

\begin{figure}
	\center
	\includegraphics[width=0.35\textwidth]{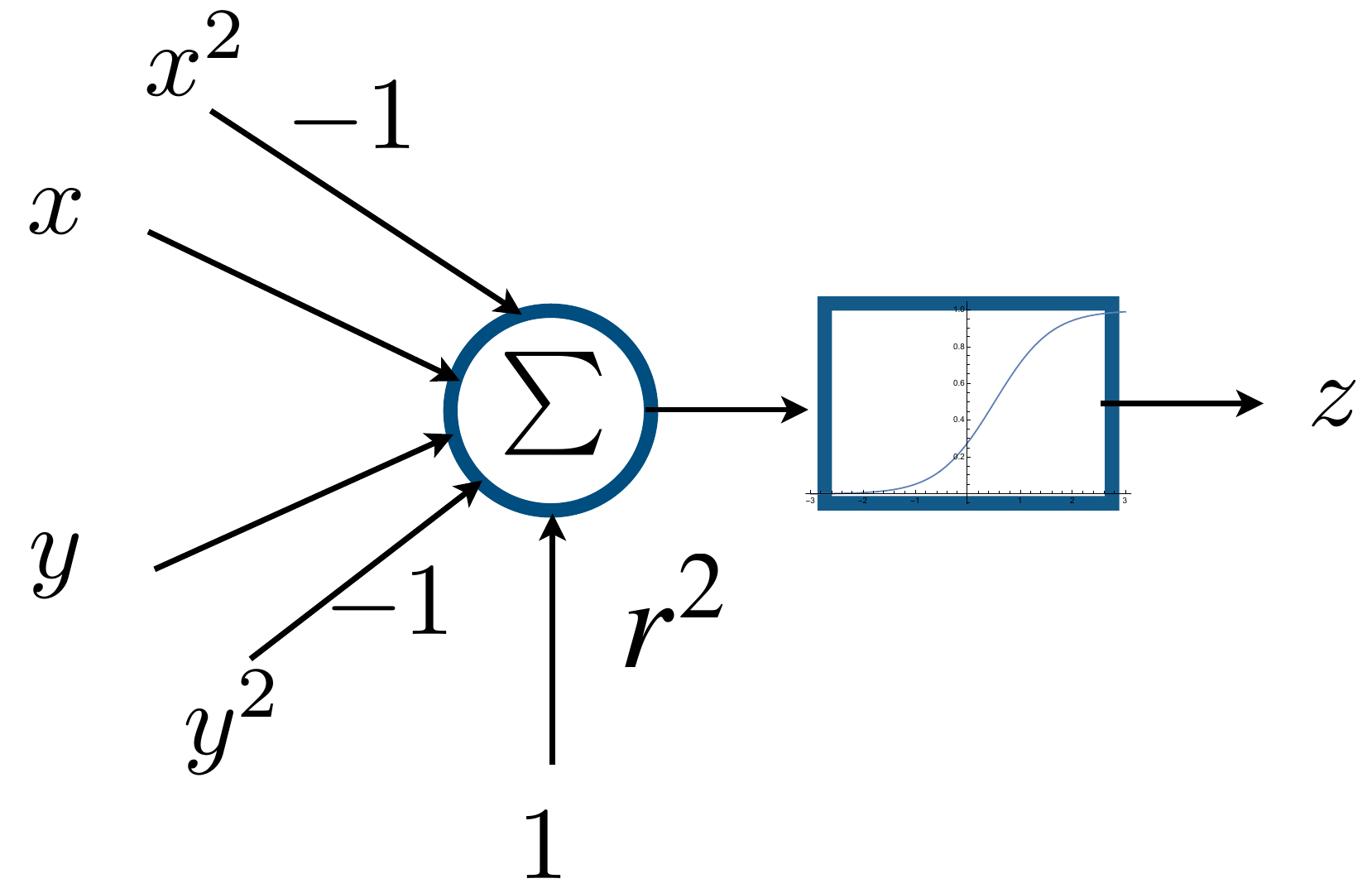}
	\caption{Perceptron model classifying a circle with radius $r$}
	\label{fig:circle}
\end{figure}

\begin{figure}
	\includegraphics[width=0.45\textwidth]{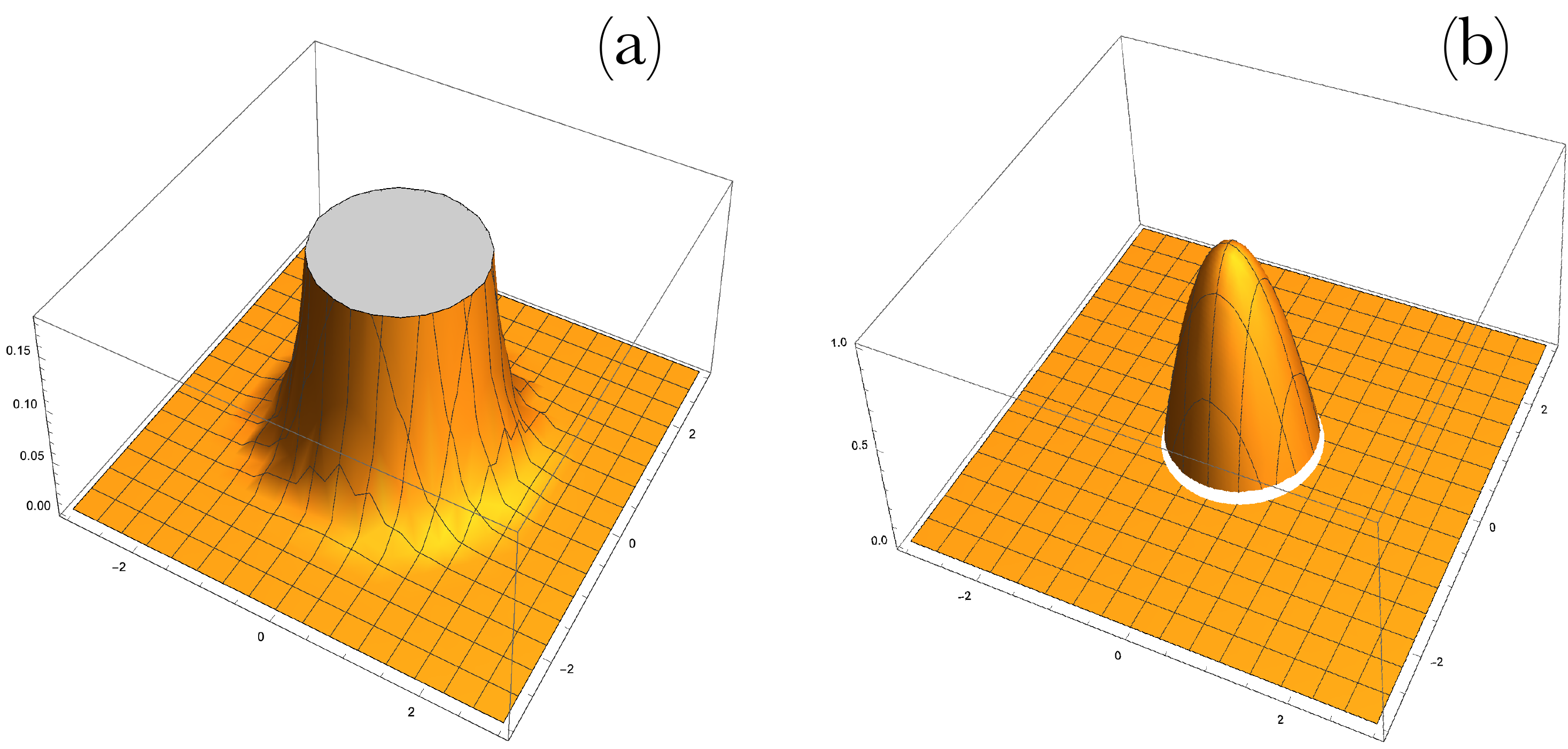}
	\caption{Output values for a circular region using nilpotent logic (a), and its differentiable approximation (b)}
	\label{fig:circle3d}	
\end{figure}

Additionally, taking into account the fact that the area inside or outside a circle is described by an inequality containing the squares of the input values, it is also possible to construct a novel type of unit by adding the square of each input into the input layer (see Figure \ref{fig:circle}). This way, the  polygon approximation of the circle can be eliminated. For an illustration, see Figure \ref{fig:circle3d}. Note that by modifying the weights, an arbitrary conic section can also be described.

Choosing the right activation function for each layer is crucial and may have a significant impact on metric scores and the training speed of the neural model. 
In the model introduced in this Section, the smooth approximation of the cutting function is a natural choice for the activation function in the first layer as well as in the hidden layers, where the logical operators work.
Although there are a vast number of activation functions (e.g. linear, sigmoid, $\tanh$, or the recently introduced Rectified Linear Unit (ReLU) \cite{relu}, exponential linear unit (ELU) \cite{elu}, sigmoid-weighted linear unit (SiLU) \cite{silu}) considered in the literature, most of them are introduced based on some desired properties, without any theoretical background. The parameters are usually fitted only on the basis of experimental results. The squashing function stands out of the other candidates by having a theoretical background thanks to the nilpotent logic which lies behind the scenes.  

To sum up, on the one hand, this structure leads to a drastic reduction in the number of parameters to be learned, and on the other hand, it supports the interpretation, making the debugging process manageable. Given the logical structure, the parameters to be learned are located in the first layer. The choice of the activation functions in the first layer as well as in the hidden, logical layers, have a sound theoretical background. 
Designing the network architecture appropriately, arbitrary regions can be described as intersections and unions of polyhedra in the decision space. Moreover, multicriteria decision tools can also be integrated with given weights and thresholds.
	Note that the weights and biases in the hidden layers define the type of operator to be used. These parameters can also be learned in a more sophisticated model to be examined in future work. 

\section{Playground Examples} \label{PE}

To illustrate our model with some simple examples, we extended the Tensorflow Playground with the squashing function ($\beta=50, \lambda=1, a=0.5$) as activation function and modified the backpropagation algorithm according to the frozen weights in the hidden layers. 
\subsection{XOR} \label{XOR}
\begin{figure}
	\begin{center}	\includegraphics[width=0.45\textwidth]{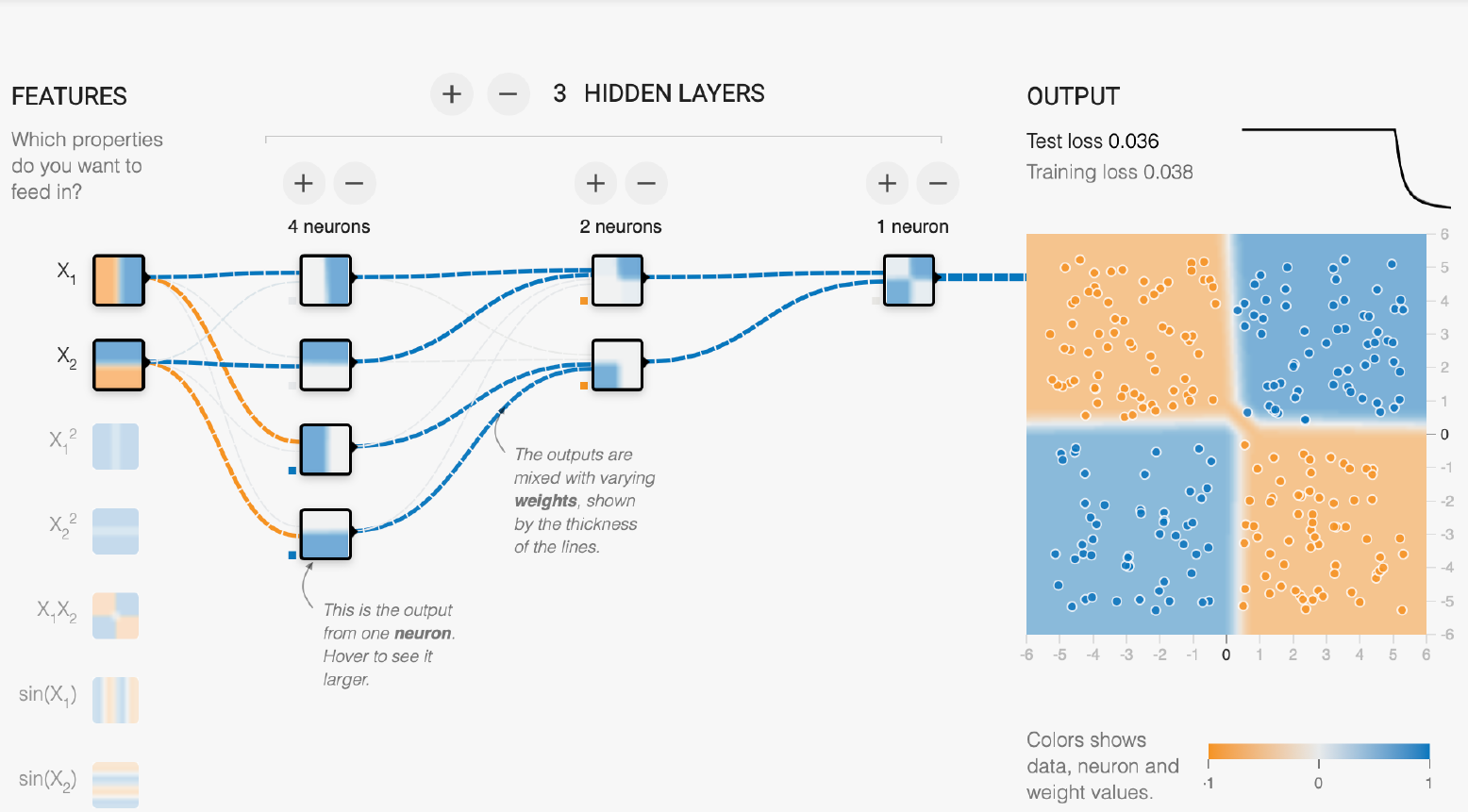}
		\caption{Nilpotent neural network block designed for modeling XOR}\label{F2}
	\end{center}
\end{figure}

Let us first consider an example on a particular data set based on Example \ref{ex:E1}. 
An image of a generated set of data is shown in Fig.~\ref{F2}. Orange data points have a value of $-1$ and blue points have a value of $+1$. 
Here, the target variable is positive when $x$ and $y$ are both positive or both negative. 
In a logical network:
\begin{itemize}
	\item If $(x_1>0)$ AND $(x_2>0)$ THEN predict $+1$
	\item If $(x_1<0)$ AND $(x_2<0)$ THEN predict $+1$
	\item Else predict $-1$
\end{itemize}




An efficient neural network can be built to make predictions for this logical expression even without using the cross feature $x*y$. For the structure and for the frozen weights and biases, see Table \ref{tab:xo}.
\begin{table*}[ht!]
	\caption{Weights and biases for modeling the XOR logical gate} \label{tab:xo}
	\begin{center}\begin{tabular}{|c|c|c|c|c|}
			\hline 
			$x>0$ & $w_{1}=1$ & $w_{2}=0$ & $b=0$ &  $[x]$ \\ 
			\hline 
			$y>0$ & $w_{1}=0$ & $w_{2}=1$ & $b=0$ &  $[y]$ \\ 
			\hline 
			$x \mbox{ AND } y$ & $w_{1}=1$& $w_{2}=1$& $b=-1$ & $[x+y-1]$ \\ 
			\hline 
			$x \mbox{ OR } y$ & $w_{1}=1$& $w_{2}=1$& $b=0$ & $[x+y]$ \\ 
			\hline 
			$\mbox{ NOT }(x)$	& $w_{1}=-1$ & $w_{2}=0$ & $b=1$ & $[1-x]$  \\ 
			\hline 
			$\mbox{ NOT }(y)$	& $w_{1}=0$ & $w_{2}=-1$ & $b=1$ & $[1-y]$ \\ 
			\hline 
			$(\mbox{ NOT }(x)) \mbox{ AND } $ & $w_{1}=-1$ & $w_{2}=-1$ & $b=1$ & $[-x-y+1]$ \\ 
		
	$(\mbox{ NOT }(y))$&&&&\\	\hline 
	\end{tabular}\end{center}
\end{table*}	 

According to our model, the smooth approximation of the cutting function called the squashing function is a natural choice for the activation function in the first layer as well as in the hidden layers, where the logical operators are used.  
If we design this logical structure before training, an interpretation of the network naturally emerges. 

Notice how the neurons in the hidden layer reveal the logical structure of the network (Figure \ref{F2}), assisting the interpretability of the neural model. 


\subsection{Preference}
\begin{figure}
	\begin{center}	\includegraphics[width=0.45\textwidth]{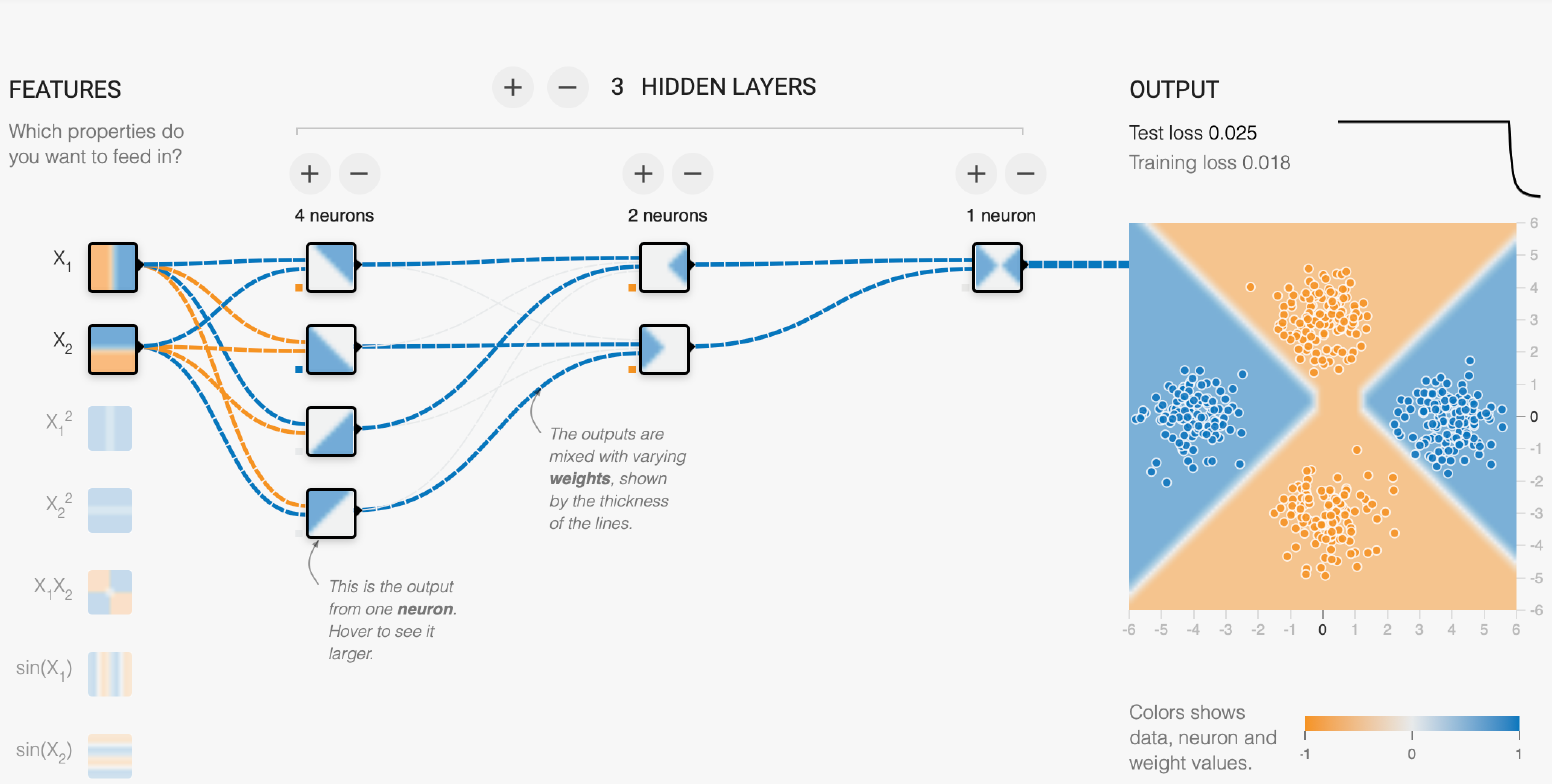}
		\caption{Nilpotent neural network block designed for modeling preference}\label{F3}
	\end{center}
\end{figure}

Another image of a generated set of data is shown in Figure \ref{F3}. Orange data points have a value of $-1$ and blue points have a value of $+1$. 
Here, the network has to learn the parameters of the straight lines separating the different regions. The target variable is positive when both $x>y$ and $y>-x$ hold or where both $x<y$ and $y<-x$ hold. In a logical network:
\begin{itemize}
	\item If $(x>y)$ AND $(y>-x)$ THEN predict $+1$
	\item If $(x<y)$ AND $(y<-x)$ THEN predict $+1$
	\item Else predict $-1$
\end{itemize}

The network structure is illustrated in Figure \ref{F3}.
\begin{table}[ht!] 
	\caption{Weights and biases for modeling the preference operator}\label{tab:rot}
	\begin{center}	\begin{tabular}{|c|c|c|c|c|}
			\hline 
			$x>y$ & $w_{1}=0.5$ & $w_{2}=-0.5$ & $b=0.5$ &  $\left[\frac{x-y+1}{2}\right]$ \\ 
			\hline 
			$y>-x$ & $w_{1}=0.5$ & $w_{2}=0.5$ & $b=0.5$ &  $\left[\frac{x+y+1}{2}\right]$ \\ 
			\hline 
			$x<y$ & $w_{1}=-0.5$ & $w_{2}=0.5$ & $b=0.5$ &  $\left[\frac{-x+y+1}{2}\right]$ \\ 
			\hline	$y<-x$ & $w_{1}=-0.5$ & $w_{2}=-0.5$ & $b=0.5$ &  $\left[\frac{-x-y+1}{2}\right]$ \\ 
			\hline 
	\end{tabular}\end{center}
\end{table}	 
Here, the expression $x>y$ is modeled by the preference operator $p(x,y)$ (see Table \ref{abc} and \ref{tab:rot}). Notice how the neurons in the hidden layer reveal the logical structure of the network (see Figure \ref{F3}), assisting the interpretability of the neural model.

\begin{figure}
	\begin{center}	\includegraphics[width=0.45\textwidth]{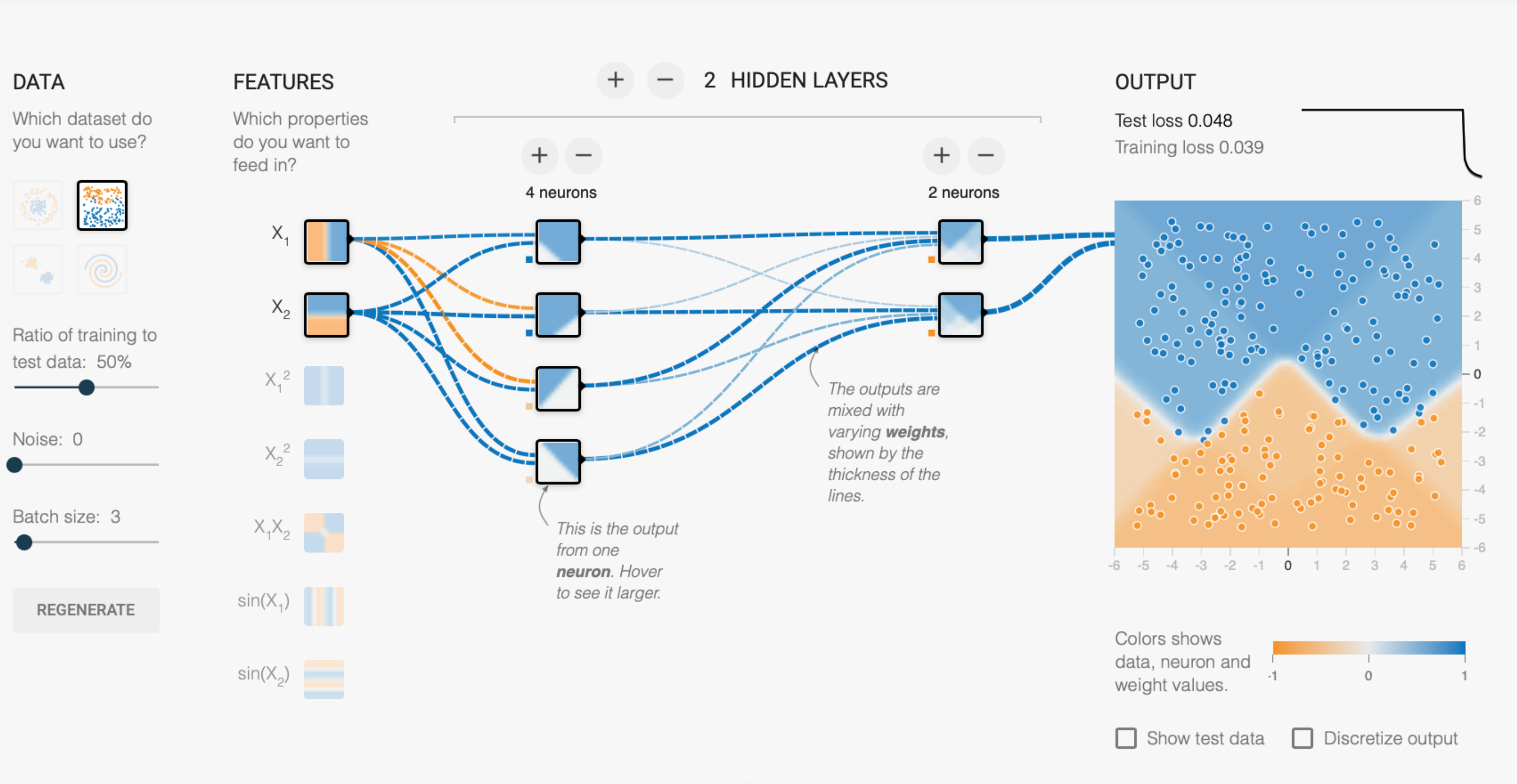}
		\caption{Nilpotent neural network block designed for finding a concave region}\label{fig:concave}
	\end{center}
\end{figure}

\begin{rem}
	Networks can also be readily designed for finding concave regions. For example, see Figure \ref{fig:concave}.
\end{rem}
\begin{rem}
	Note that the frequently used $\min$ and $\max$ operators can also be modeled by a similar network, based on Equations \eqref{min} and \eqref{max}.
\end{rem}

\vspace{-0.1cm}
\section{Conclusions}\label{concl}
In this study, we suggested interpreting neural networks by using continuous nilpotent logic and multicriteria decision tools to reduce the black box nature of the neural models, aiming at the interpretability and improved safety of machine learning. We introduced the main concept and the basic building blocks of the model to lay the foundations for the future steps of the application. In our model, membership functions (representing truth values of inequalities), and also nilpotent operators are modeled by perceptrons. The network architecture is designed prior to training. In the first layer, the parameters of the membership functions are needed to be learnt, while in the hidden layers, the nilpotent logical operators work with given weights and biases. Based on previous results, a rich asset of logical operators with rigorously examined properties is available. A novel type of neural unit was also introduced by adding the square of each input to the input layer (see Figure \ref{fig:circle}) to describe the inside or the outside of a circle without polygon approximation.

The theoretical basis offers a straightforward choice of activation functions: the cutting function or its differentiable approximation, the squashing function. Both functions represent truth values of soft inequalities, and the parameters have a semantic meaning. Our model also seems to provide an explanation to the great success of the rectified linear unit (ReLU). 

The concept was illustrated with some toy examples taken from an extended version of the tensorflow playground.
The implementation of this hybrid model in deeper networks (by combining the building blocks introduced here) and its application e.g. in multicriteria decision making or image classification is left for future work.

\vspace{-0.19cm}
\section*{Acknowledgments}
This study was partially supported by grant TUDFO\-/47138-1/2019-ITM of the Ministry 
for Innovation and Technology, Hungary.

\section*{References}

\bibliographystyle{elsarticle-num}

\end{document}